\documentclass[conference,english]{IEEEtran}
\IEEEoverridecommandlockouts
\usepackage{amsmath,amssymb,amsfonts}
\usepackage{algorithmic}
\usepackage{graphicx}
\usepackage{textcomp}
\usepackage[table]{xcolor}

\usepackage[utf8]{inputenc}
\usepackage{babel,csquotes,xpatch}

\usepackage[style=ieee,backend=biber,useprefix=true]{biblatex}
\usepackage{url}
\usepackage{makecell}
\usepackage{booktabs}

\newcommand{\hb}[1]{\cellcolor{black!20}{#1}}
\newcommand{\ha}[1]{\cellcolor{black!20}\textbf{#1}}

\usepackage[acronym]{glossaries}
\newacronym{ES-RNN}{ES-RNN}{Exponential Smoothing - Recurrent Neural Network}
\newacronym{FFORMA}{FFORMA}{Feature-based FORecast Model Averaging}
\newacronym{N-BEATS}{N-BEATS}{Neural Basis Expansion Analysis}
\newacronym{RF}{RF}{Random Forests}
\newacronym{XGBoost}{XGBoost}{eXtreme Gradient Boosting Machine}
\newacronym{MLP}{MLP}{Multilayer Perceptron}
\newacronym{ANN}{ANN}{Artificial Neural Network}
\newacronym{RNN}{RNN}{Recurrent Neural Network}
\newacronym{LSTM}{LSTM}{Long Short Term Memory}
\newacronym{ARIMA}{ARIMA}{AutoRegressive Integrated Moving Average}
\newacronym{ML}{ML}{Machine Learning}
\newacronym{OWA}{OWA}{Overall Weighted Average}
\newacronym{sMAPE}{sMAPE}{Symmetric Mean Absolute Percentage Error}
\newacronym{MAPE}{MAPE}{Mean Absolute Percentage Error}
\newacronym{MASE}{MASE}{Mean Absolute Scaled Error}
\newacronym{ES}{ES}{Exponential Smoothing}
\newacronym{FFORMS}{FFORMS}{Feature-based FORecast Model Selection using Random Forest}
\newacronym{FFORMS-G}{FFORMS-G}{Feature-based FORecast Model Selection using Gradient Boosting}
\newacronym{FFORMA-N}{FFORMA-N}{Feature-based FORecast Model Averaging using Neural Networks}
\newacronym{DeFORMA}{DeFORMA}{Deep-learning FORecast Model Averaging}
\newacronym{NN-STACK}{NN-STACK}{Neural Networks Stacking Regression}
\newacronym{ReLU}{ReLU}{Rectified Linear Unit}
\newacronym{AVG}{AVG}{Simple Model Averaging}
\newacronym{SOTA}{SOTA}{state-of-the-art}

\begin{document}

\title{Late Meta-learning Fusion Using Representation Learning for Time Series Forecasting
\thanks{We acknowledge the funding support of the Nedbank Research Chair}
}

\author{\IEEEauthorblockN{1\textsuperscript{st} Terence L. van Zyl}
\IEEEauthorblockA{\textit{Institute for Intelligent Systems} \\
\textit{University of Johannesburg}\\
Johannesburg, South Africa \\
Contact: tvanzyl@gmail.com  0000-0003-4281-630X}
}

\maketitle

\begin{abstract}
Meta-learning, decision fusion, hybrid models, and representation learning are topics of investigation with significant traction in time-series forecasting research. Of these two specific areas have shown \acrlong{SOTA} results in forecasting: hybrid meta-learning models such as \gls{ES-RNN} and \gls{N-BEATS} and feature-based stacking ensembles such as \gls{FFORMA}. However, a unified taxonomy for model fusion and an empirical comparison of these hybrid and feature-based stacking ensemble approaches is still missing. This study presents a unified taxonomy encompassing these topic areas. Furthermore, the study empirically evaluates several model fusion approaches and a novel combination of hybrid and feature stacking algorithms called \gls{DeFORMA}. The taxonomy contextualises the considered methods.
Furthermore, the empirical analysis of the results shows that the proposed model, \gls{DeFORMA}, can achieve \acrlong{SOTA} results in the M4 data set.  \gls{DeFORMA}, increases the mean \gls{OWA} in the daily, weekly and yearly subsets with competitive results in the hourly, monthly and quarterly subsets. The taxonomy and empirical results lead us to argue that significant progress is still to be made by continuing to explore the intersection of these research areas.
\end{abstract}


\glsresetall
\section{Introduction}

Given a dependent variable, numerous forecasts can be generated from models with different structural assumptions or initialisations. Since the no-free-lunch theorem holds, no single forecasting model would universally outperform all others~\cite{wolpert1997no}. A crucial question emerges for selecting the best approach for a specific time series. Better outcomes can almost always be achieved by combining predictions rather than making a single selection~\cite{bishop2006pattern, reid1972comparison}. Further, Research provides substantive evidence that this fusion of predictive models also applies to time series forecasting methods~\cite{reich2019collaborative, mcgowan2019collaborative, johansson2019open, cawood2022evaluating}. 

An approach to the fusion of forecasting models that has stood out is \gls{FFORMA}, which is based on the features of the time series extracted using third-party instruments. A meta-learner then uses the extracted features to select an appropriately weighted combination of heterogeneous base models~\cite{montero2020fforma, cawood2022evaluating}. Despite the success of \gls{FFORMA} and variants of their off, they require preprocessing of the time series to extract features. Further, these features are hand-engineered, limiting the knowledge that can be extracted from the time series. Similar limitations were typical in the early hand-engineering filter approaches to computer vision. Representation learning has improved the \gls{SOTA} in numerous machine learning fields, especially in computer vision and natural language. With the successes in these domains has come an exploration of similar deep learning methods in time series forecasting. The rapid exploration means the taxonomy for forecasting fusion still needs to be completed, making it difficult to fully contextualise new approaches when encountered in the literature~\cite{cawood2022evaluating}.

These two open areas require further evaluation. First, to what extent might the use of representation learn as a drop-in replacement for hand-engineered time series features lead to improvements in time series forecasting? Second, can re-examining time series forecasting studies involving model fusion provide a complete taxonomy? To this end, the contributions of this paper are three-fold:
\begin{itemize}
    \item First, the study contributes towards a complete taxonomy of decision fusion, specifically for forecasting.
    \item Second, the study demonstrates \gls{SOTA} results for the M4 dataset using representation learning as a replacement for hand-engineered features in the context of forecasting fusion.
    \item Third, the study presents two novel temporal heads supporting the results of \gls{SOTA}.
\end{itemize}

\section{Forecast Model Fusion: Taxonomy}

\begin{figure*}
    \centering
    \includegraphics[width=\textwidth]{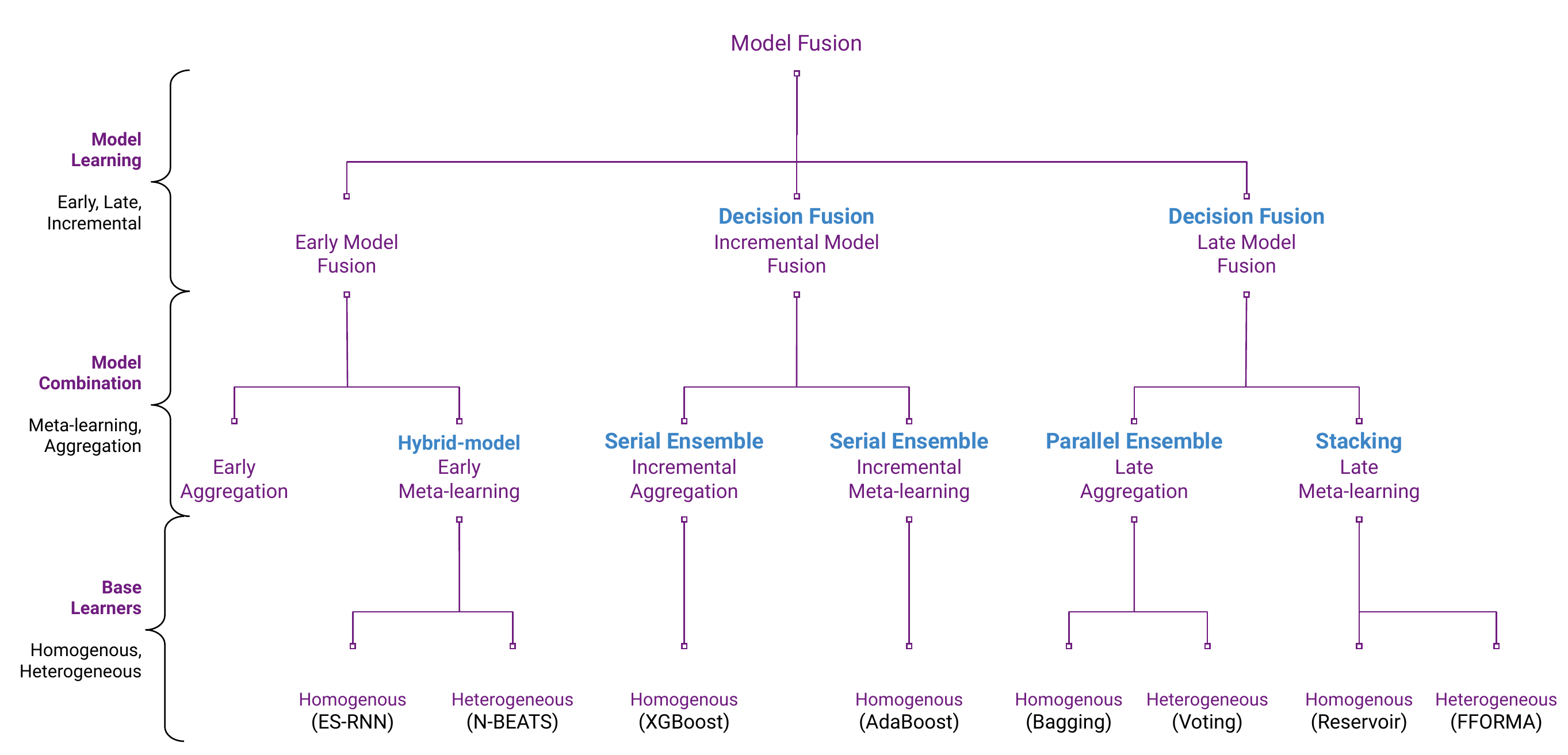}
    \caption{An taxonomy of model fusion. Showing Decision Fusion, Ensemble Learning, Stacking, and Hybrid-models, among others~\cite{ribeiro2020ensemble,zhang2003time}.}
    \label{fig:model_fusion}
\end{figure*}


This section examines meta-learning, model fusion and decision fusion as related concepts. It expands upon an existing taxonomy that previously jointly contextualised them~\cite{cawood2022evaluating}—first, a description of meta-learning within the context of time series forecasting. Next, decision fusion is described as it relates to forecasting and how it differs from traditional data fusion. Finally, ideas from the three areas; decision fusion, data fusion, and meta-learning; extended the existing taxonomy for model fusion. The expanded taxonomy allows us to analyse the proposed forecasting fusion techniques described in related work at the end of this section.

Consider models parameterised by data to improve performance on a task. These models are typical of machine learning, where an internal learning model is trained using data and a learning objective during base-model learning. These models are ``base-learners''~\cite{vilalta2004using, wang2009rule, wang2009rule}. ``Meta-learning'' refers to techniques that learn a relationship between ``meta-knowledge'' (data and task characteristics) and the performance of the underlying base-learners. During meta-learning, an outer algorithm uses meta-knowledge to update the inner learning algorithm to improve the model on an outer learning objective~\cite{hospedales2021meta}. In the context of time series, this process automatically acquires knowledge to select or fuse time series forecast models~\cite{prudencio2005using, lemke2010meta}.

Coincident with the above discussion, the word ``fusion'' refers to integrating information or knowledge from multiple sources. Fusion can, in some instances, be split into several sub-types: three of which are data fusion, feature fusion, and decision fusion~\cite{dasarathy1997sensor}. Here the fusion approaches are classified according to the processing level at which the fusion occurs. These processing levels are termed early fusion or data-level fusion, late fusion or decision-level fusion, and intermediate fusion or feature-level fusion that combines late and early approaches~\cite{michelsanti2021overview}.

Of these processing-level categories of fusion, decision fusion is aimed at learning to combine the beliefs of the collection of models into a single consensus belief~\cite{sinha2008estimation}. Alternatively, one might say that decision fusion integrates the ``decisions'' of several base-learners into a single ``decision'' about a target~\cite{zhang2022decision}. This approach to ``combining the wisdom of crowds'' is also sometimes termed ensemble learning~\cite{goodfellow2016deep} and, in some studies, has been called ``fusion models''~\cite{nalabala2021financial,thakkar2021fusion}. However, this view of fusion now excludes areas such as hybrid models and deep learning covered by meta-learning~\cite{schaul2010metalearning}. From a processing-level viewpoint of fusion, they are now categorised outside the boundaries of decision fusion.

Approaches with a processing-level view of fusion look to combine multi-modal, multi-resolution, and multi-temporal sources to produce more consistent, accurate, and valuable higher-level products~\cite{sinha2008estimation}. Analogously, we are now positioned to take a learning view of fusion. Let us consider the definition of~\textcite{cawood2022evaluating} for model fusion: ``\emph{the integration of base-learners to produce a lower biased and variance; and a more robust, consistent, and accurate meta-model}.'' The presented taxonomy now sub-categorises model fusion along an axis based on whether the learning process is early, incremental, or late, with them being described as~\cite{cawood2022evaluating}:
\begin{LaTeXdescription}
    \item[\textbf{Early model fusion}] integrates the base-learners before they have been parameterised (trained), and the combined model is trained as a single fused model.
    \item[\textbf{Late model fusion}] first parameterises (pretrains) the base-learners individually. The base-learners are integrated into a single model without further modification. 
    \item[\textbf{Incremental model fusion}] performs model integration while parameterising (training) the base-learners one at a time. The already integrated base-learners' parameters remain fixed once trained. However, the base-learner being incrementally added is trained during the integration step.
\end{LaTeXdescription}
The choice of the word ``model'' rather than ``decision'' is attributed to the fact that in the case of early model fusion, no decisions exist at the time of fusion. Decision fusion refers to only the incremental and late model fusion cases. One might argue that the correct term for early fusion is feature or data fusion. These terms are also valid but take a different processing-level view.

In the taxonomy for decision-fusion combiners by \textcite{zhang2022decision}, we note that algorithms can be grouped into two distinct groups. The first group are those methods that use a simple aggregation scheme for combination. The second group is those that use more complex meta-learning. These groupings of combiners have previously been referred to by~\textcite{ponti2011combining} as fixed or trained combiners. We are now in a position to explain the model-combining process of the taxonomy:
\begin{LaTeXdescription}
    \item[\textbf{Meta-learning fusion}] uses meta-learning to perform the model integration process (Trained Combiner). 
    \item[\textbf{Elementary fusion}] \cite{ponti2011combining}, or \textbf{Aggregation} for short, uses a simple aggregation scheme, like averaging or voting, to perform the model integration (Fixed Combiner).
\end{LaTeXdescription}

Finally, the nature of the base-learners needs to be considered. base-learners are sub-categorised as \textbf{homogeneous} if they come from the same hypothesis class (e.g. decision trees in a random forest or perceptrons in a neural network). If the base-learners come from different hypothesis classes (e.g. neural network and support vector machine), we would refer to them as \textbf{heterogeneous}~\cite{zhang2022decision}. 

Presenting the above discussion with some of the common pseudonyms for the classes of techniques allows us to arrive at the \textit{more complete} taxonomy for model fusion shown in Figure~\ref{fig:model_fusion}. The following section reviews the relevant literature for forecasting model fusion in the presented taxonomy's context.

\subsection{Related Work}

One of the earliest examples of forecasting model fusion is by~\textcite{arinze1994selecting}. They introduce stacking as a late meta-learning fusion technique that selects one forecasting model among several based on the learnt performance for six-time series meta-features. \textcite{raftery1997bayesian} suggested that choosing a single model creates issues of model uncertainty and proposed using Bayesian model averaging instead. More recently, a more robust approach has been to set the weights using the predictions of heterogeneous base-learners using cross-validation instead of posterior probabilities~\cite{clarke2003comparing}. \textcite{cawood2021feature} proposed increasing meta-features in the stacking approach by including recent statistics, such as linearity, curvature, stability, and entropy~\cite{lorena2018data, barak2019time, montero2020fforma}.

\textcite{ribeiro2020ensemble} studied late and incremental model fusion in forecasting time series. They compare the bagging, gradient-boosting, and stacking of base-learners. Their results suggest that boosting methods generally produce the lowest prediction errors. The top two submissions to the M4 forecasting competition supported these findings, also implementing ensemble learning. 

The runner-up in the M4 Competition, \gls{FFORMA}, uses late meta-learning fusion with heterogeneous base-learners. The stacking is implemented using extreme gradient boosting as the meta-learner. The meta-learner learns the weightings among several base-learners using meta-features extracted from the input time series~\cite{montero2020fforma}. The weightings learnt are then used to combine the models' predictions as a feature-weighted average.

Hybrid models use early meta-learning with heterogeneous base-learners to achieve model fusion. They are the preferred family of fusion models and have retained traction in recent time series forecasting research \cite{liu2014hybrid, wang2015study, qin2019hybrid, shinde2020forecasting}. Hybrid models were initially proposed by Zhang~\cite{zhang2003time} for forecasting, who showed that the fusion of the \gls{ARIMA} and \gls{MLP} models produces improved accuracy. Numerous hybrids implement \glspl{ANN} or combine them with traditional models~\cite{mathonsi2020prediction, laher2021deep, mathonsi2022statistics}. 

The \gls{ES-RNN} is a hybrid forecasting model and winning submission of the M4 Competition. \gls{ES-RNN} fuses a modified Holt-Winters and dilated \gls{LSTM} stacks~\cite{ESRNN} in early meta-learning using heterogeneous base-learners. As it stands, \gls{N-BEATS} is a hybrid model that achieves \gls{SOTA} forecasting accuracy by 3\% over the \gls{ES-RNN}~\cite{oreshkin2019n}. \gls{N-BEATS} is a hybrid forecasting method that integrates a fully connected neural network with traditional time series decomposition. 

\section{Deep-learning FORecast Model Averaging (DeFORMA)}

\begin{figure}
    \centering
    \includegraphics[width=\columnwidth]{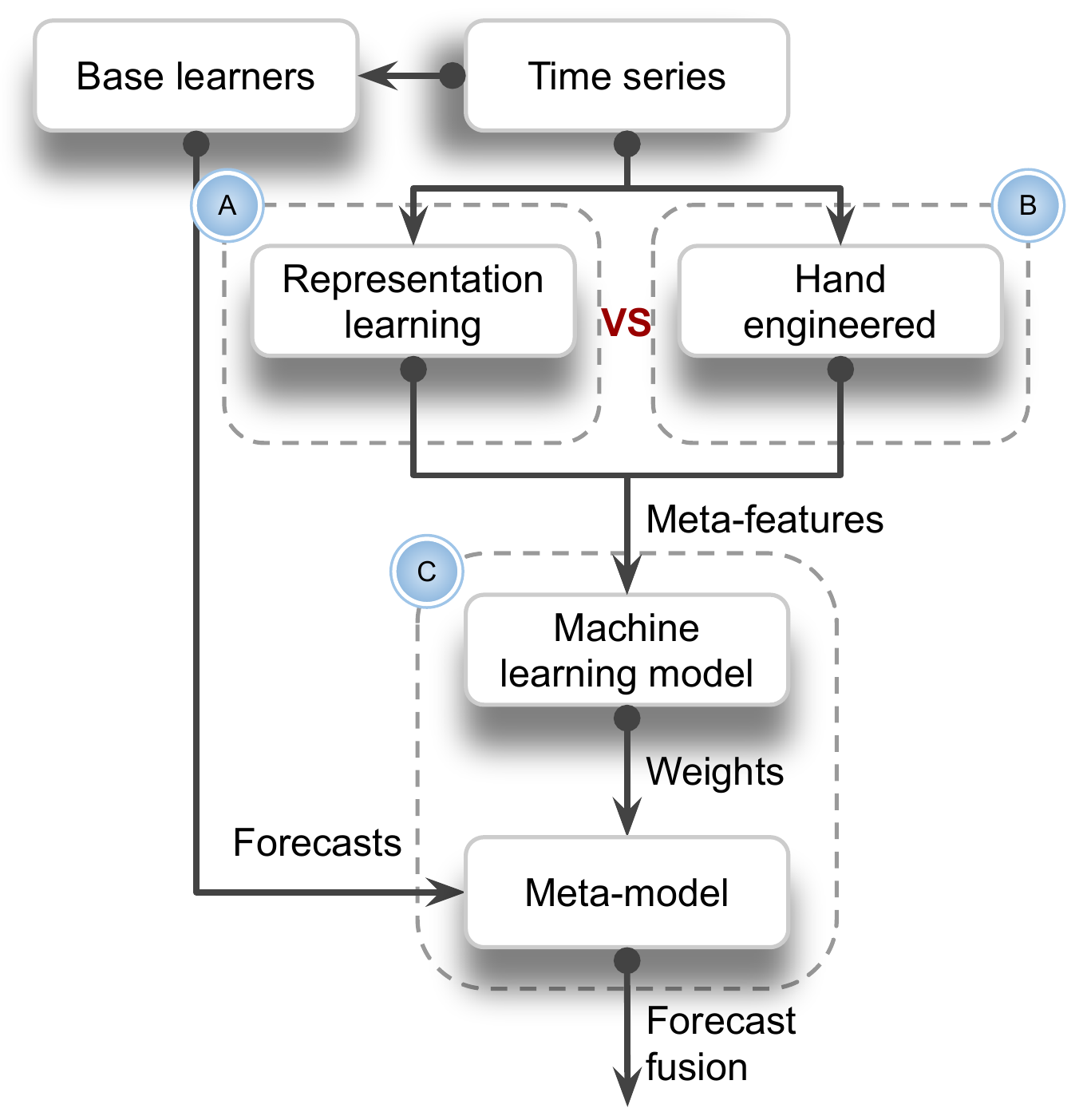}
    \caption{(A) \gls{DeFORMA} versus (B) \gls{FFORMA} approaches with respect to the (C) meta-learning. For \gls{DeFORMA}, the Machine learning model is a fully connected neural network layer. For \gls{FFORMA}, the Machine learning model is Gradient Boosting.}
    \label{fig:fformabsdeforma}
\end{figure}

Much of the success of the hybrid models (\gls{ES-RNN},\gls{N-BEATS}) stems from their ability to learn feature representations and use those representations in the model fusion process. Alternatively, \gls{FFORMA} uses gradient boosting with the hand-engineered meta-features as input to weight heterogeneous base-learners' forecasts. A natural question arises about the efficacy of using representation learning in a late meta-learning model such as \gls{FFORMA}.

\gls{DeFORMA} like \gls{FFORMA} is a late meta-learning fusion method with heterogeneous base-learners. The distinction, seen in Figure~\ref{fig:fformabsdeforma}, is that instead of using hand-engineered meta-features \gls{DeFORMA} uses representing learning to learn meta-features from the data. To accomplish this, \gls{DeFORMA} uses the same loss function
\begin{equation*}
    \sum_i^\mathrm{\#Base-learners} W_i F_i
\end{equation*}
as \gls{FFORMA}, where $W_i$ are the weights learnt for each base-learner forecast $F_i$. \gls{DeFORMA} replaces the gradient boosting model with a deep learning model. The remainder of this section discusses the neural network architecture and the resulting hyper-parameter choices.

\subsection{Neural Network Architecture}

\begin{figure}[!hbt]
    \centering
    \includegraphics[width=\columnwidth]{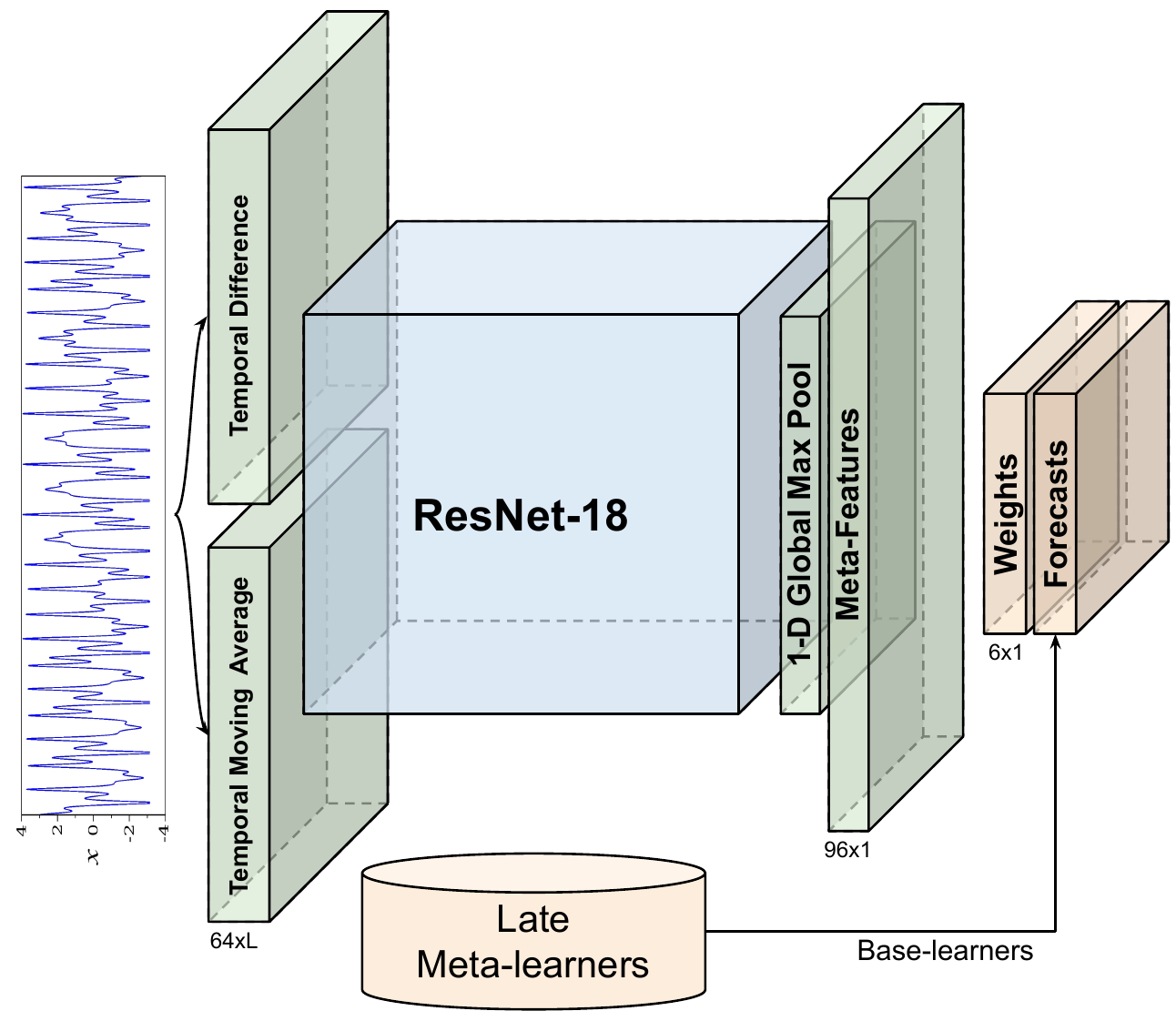}
    \caption{The \gls{DeFORMA} architecture showing the modifications to the ResNet-18 backbone in green. Orange shows the components of the \gls{FFORMA} loss.} 
    \label{fig:deforma_architecture}
\end{figure}

At its core, \gls{DeFORMA} uses a one-dimensional (1-D) ResNet-18 as the backbone, as seen in Figure~\ref{fig:deforma_architecture}. Researchers initially designed ResNet for images. Consequently, certain modifications are required. Firstly, all convolutions are replaced with 1D convolutions, and the global average pooling layer is replaced with a 1D global max pooling layer. Secondly, two Temporal heads are prepended to the input layer of the ResNet. The first Temporal head is a Differencing head. The second is a Moving Average head. More details descriptions of these Temporal heads are presented at the end of this sub-section. Finally, for short time series, the max pooling and two strides within the ResNet blocks for a deep neural network result in filters with a length of less than one. For this reason, the architecture limits the number of ``halvings'' (stride 2) that the ResNet blocks take.

The resulting architecture has the following hyper-parameters that require tuning: i) the number of halvings, ii) the number of convolutional filters, iii) the number of features in the output layer, iv) the maximum length of the time series and v) the dropout rate.

\subsubsection{Temporal Differencing Head}

The Temporal Differencing head has as its primary purpose to remove seasonal trends. To achieve this, the head consists of three layers. The first layer is a 1-D convolution layer ($F \times L$) consisting of $F$ filters and having length $L$ (no bias). The length $L$ of the filters equals the seasonality$S+1$ of the time series. The weights of filters are constrained such that they must \textit{sum to zero}. This convolution is followed by Layer Normalisation and then 1D Spatial Dropout.

\subsubsection{Temporal Moving Average Head}

The Temporal Moving Average head has as its primary purpose to remove seasonality. To achieve this, the head consists of three layers. The first layer is a 1-D convolution layer ($F \times L$) consisting of $F$ filters and having length $L$ (no bias). The length $L$ of the filters is equal to the seasonality$S$ of the time series. The weights of filters are constrained such that they must \textit{sum to one}. This convolution is followed by Layer Normalisation and then 1D Spatial Dropout.

\section{Method}

To test the primary hypothesis that learning the features required for \gls{FFORMA} would be possible, the study empirically evaluated numerous \gls{SOTA} models against the proposed solution. The accuracy of the forecast for this evaluation is assessed by measuring the mean and median \gls{OWA} on the M4 dataset. To this end, our experiment follows the procedure used by \textcite{cawood2022evaluating}.

\subsection{Dataset}

\begin{table}[tb]
\centering
\caption{Number of series per data frequency and domain \cite{makridakis2020m4}.}
\resizebox{\columnwidth}{!}{
\begin{tabular}{l|rrrrrrr}
    Data subset 
    & Micro
    & Industry
    & Macro
    & Finance
    & \makecell[b]{Demo-\\graphic}
    & Other
    & Total\\ \bottomrule\toprule
Yearly (Y)    & 6,538  & 3,716  & 3,903  & 6,519  & 1,088 & 1,236 & 23,000  \\ 
Quarterly (Q) & 6,020  & 4,637  & 5,315  & 5,305  & 1,858 & 865   & 24,000  \\ 
Monthly (M)   & 10,975 & 10,017 & 10,016 & 10,987 & 5,728 & 277   & 48,000  \\ 
Weekly (W)    & 112    & 6      & 41     & 164    & 24    & 12    & 359  \\ 
Daily (D)     & 1,476  & 422    & 127    & 1,559  & 10    & 633   & 4,227  \\ 
Hourly (H)    & 0      & 0      & 0      & 0      & 0     & 414   & 414  \\ 
Total         & 25,121 & 18,798 & 19,402 & 24,534 & 8,708 & 3,437 & 100,000  \\
\bottomrule
\end{tabular}
}
\label{tab:m4data}
\end{table}

Makridakis Competitions (or M-Competitions) establish an inventory of standard benchmark data sets that are the most convincing effort driving continuous advancement in forecasting techniques~\cite{makridakis2020forecasting, makridakis2021m5}. The recent M4 Competition conducted a large-scale comparative study of $61$ forecasting methods. The comparison included statistical methods, for example, \gls{ARIMA} and Holt-Winters, and machine learning methods, for example, \gls{MLP} and \gls{RNN}. Similar to previous studies~\cite{makridakis2020m4}, they find that employing an incremental and late model fusion of base-learners delivers the most promising results~\cite{reich2019collaborative, mcgowan2019collaborative, johansson2019open}. Before the M5 competition, \textcite{makridakis2020forecasting} suggested that machine learning models showed inferior forecasting performance overall. More recently, the M5 Competition was the first in the series with hierarchical time series data~\cite{makridakis2021m5}. The M5 Competition confirmed late meta-learning fusion's dominance in forecasting time series data. The top submissions were techniques that adopted parallel ensemble learning and gradient boosting, often with deep learning models as base-learners~\cite{makridakis2022m5}. However, two early meta-learning hybrid models that use deep learning stand out, both of which have been \gls{SOTA} results for M4: \gls{ES-RNN} and \gls{N-BEATS}.

\textcite{cawood2022evaluating} demonstrated that the results of the M5 competition are valid for the M4 competition. For this reason, this study focuses on the M4 Competition. The M4 dataset contains a $100,000$ time series of different seasonalities. The minimum time series length is as low as $13$ for the yearly series and $16$ for quarterly. Data are available on Github\footnote{https://github.com/Mcompetitions/M4-methods/tree/master/Dataset}, and Table~\ref{tab:m4data} provides a summary of the number of series per frequency and domain. Subset names H, D, W, M, Y and Q of Tables~\ref{tab:m4data} - \ref{tab::owa_results_median} correspond to the hourly, daily, weekly, monthly, yearly and quarterly seasonality subsets. Domains include economics, finance, demographics, industries, tourism, trade, labour and wages, real estate, transportation, natural resources and the environment. 

\subsubsection{Data Preprocessing}

Little additional preprocessing is applied other than limiting the maximum length of input series as a requirement from TensorFlow. Further, if the time series is shorter than the maximum length, then zero padding is applied. The maximum length is treated as a hyperparameter. However, the ablation study shows that the proposed method is insensitive to this value.

\subsection{Base-Learners}
\label{section::stat_base_models}

This section describes the base-learners used.

\subsubsection{Auto-ARIMA} a classic approach for benchmarking forecast methods' implementations. This study uses the forecasts from an Auto-ARIMA method that uses the maximum likelihood estimation to approximate the parameters \cite{hyndman2008automatic}.    
\subsubsection{Comb (or COMB S-H-D)} is the arithmetic average of three exponential smoothing methods: Single, Holt-Winters, and Damped exponential smoothing \cite{makridakis2000m3}. Comb was the winning approach for the M2 competition and was used as a benchmark in the M4 competition.
\subsubsection{Theta} the most promising approach of the M3 competition \cite{assimakopoulos2000theta}. Theta is a straightforward forecasting method that averages the extrapolated Theta lines, computed from two given Theta coefficients, applied to the second differences of the time series.    
\subsubsection{Damped  Holt's} is exponential smoothing with a trend component modified with a damping parameter $\phi$ imposed on the trend component~\cite{holt2004forecasting,mckenzie2010damped}.
\subsubsection{\gls{ES-RNN}} is used as a base-learner and is described in more detail below~\cite{ESRNN}.

\subsection{Compared Forecasting Methods}

This section describes the compared fusion models.

\subsubsection{\gls{ES-RNN}~\cite{ESRNN}} is a hybrid forecasting method. It produces more accurate forecasts using an early meta-learning fusion of homogeneous \gls{ES} base-learners with a \gls{LSTM} meta-learner. \gls{ES-RNN} was the top performing model of the M4 competition.

\subsubsection{\gls{FFORMA}~\cite{montero2020fforma}} adopts a feature-weighted model averaging strategy. A meta-learner, gradient boosting~\cite{chen2016xgboost}, learns to weight the effectiveness of a base-learner pool for different regions of a set of hand-engineered meta-features. The meta-learner then uses a late fusion of the predictions of the base-learners based on the learned weightings.

\subsubsection{\gls{FFORMS}~\cite{talagala2018meta}} was the precursor to \gls{FFORMA} and used a random forest~\cite{liaw2002classification} for late meta-learning to select a single model from a pool of heterogeneous base-learners. The selection is based on their varying performance observed over the meta-data feature space. Following the original paper, the study used Gini impurity and the same meta-features as in \gls{FFORMA}.


\subsubsection{\gls{FFORMA-N}~\cite{cawood2022evaluating}} uses a \gls{MLP} for late meta-learning that takes the same meta-features as \gls{FFORMA} as inputs and a one-hot encoded vector of the best-performing model as the model's targets.

\subsubsection{\gls{N-BEATS}~\cite{oreshkin2019n}} is a hybrid forecasting method that uses a \gls{MLP} as a meta-learner with traditional time series decomposition blocks as base-learners. 

\subsubsection{\gls{AVG}} uses a simple model averaging as a baseline for comparison.

\subsection{Experimental Setup}

\begin{figure*}
    \centering
    \includegraphics[width=0.7\textwidth]{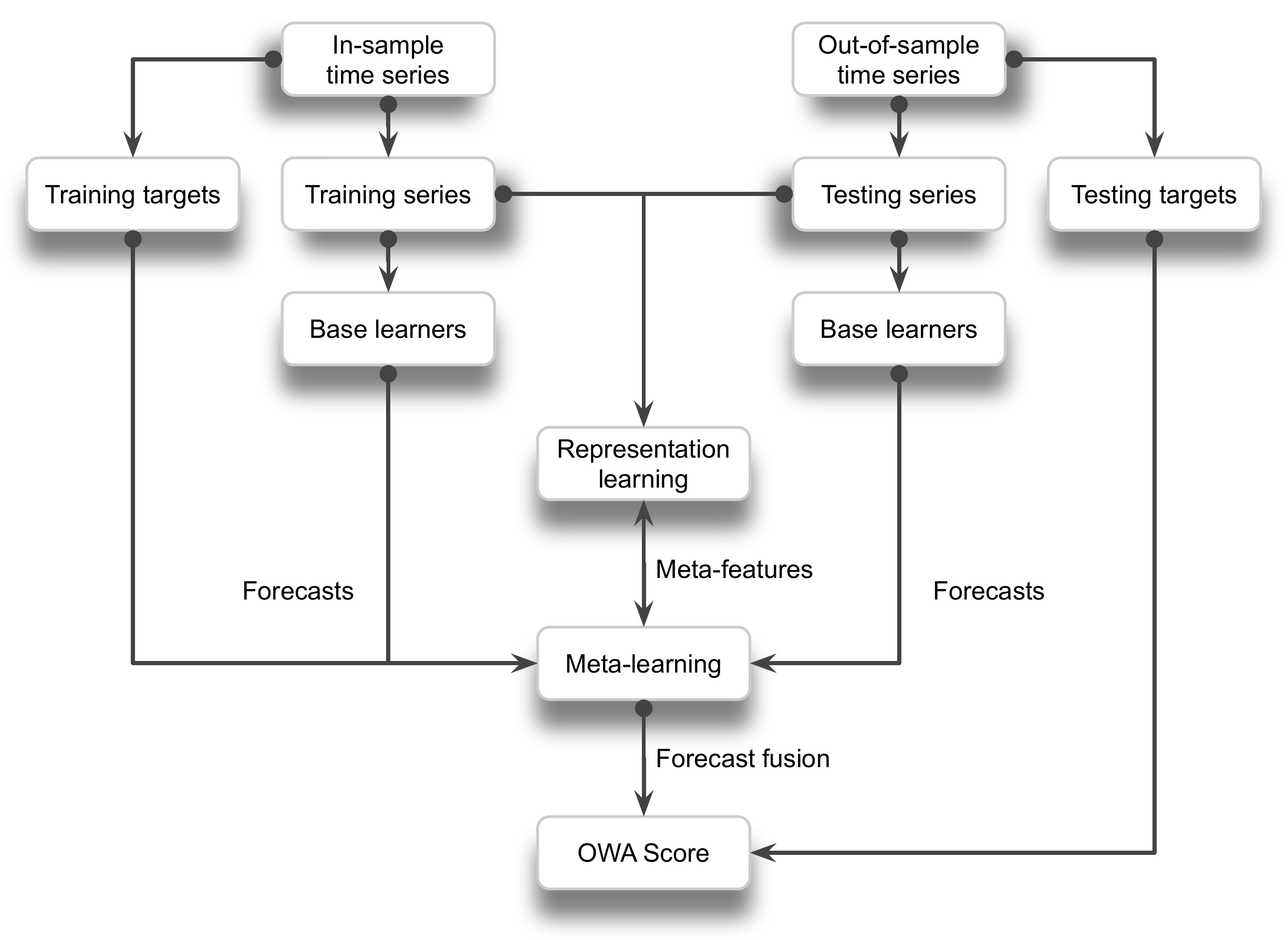}
    \caption{The proposed methodology integrates representation learning into the machine learning pipeline.}
    \label{fig:method}
\end{figure*}

All comparisons are made using ten-fold cross-validation. Cross-validation is repeated five times, and scores are averaged over all 50 runs to rule out false positive results from random fluctuations. A pseudo-random number generator \cite{blum1986simple} is configured to produce the indices to split the data into training and validation sets for each of the five runs.

Figure~\ref{fig:method} shows a more detailed view of the proposed methodology. The training and test splits are as per the M4 dataset. Base-learner forecasts are acquired from the M4 competition. The length of the training and testing target forecast horizons also follow the original M4 dataset. 

\subsubsection{Hyper-parameter Tuning}

The architectures and hyper-parameters were all found using only the training set of the first fold of the first cross-validation run. Since our training and testing setup follows from ~\textcite{cawood2022evaluating}, the experiments in this study employ the hyper-parameters found in that study for \gls{FFORMA} and its variants. The neural networks were tuned using backtesting, and the gradient boosting methods were tuned using Bayesian optimisation. 

For \gls{DeFORMA}, the architectures and hyper-parameters were also found using only the training set of the first fold of the first cross-validation run. The hyper-parameters were determined using a course grid search. 

For model architecture, the hyper-parameters were selected from: i) Halvings - $[1, 2, 3, 4, 5]$, ii) Convolutional filters - $[64]$ iii) Meta-features - $[32, 40, 64, 96, 128]$ iv) Maximum length - $[32, 64, 96, 128, 160, 192, 224, 256, 288, 320, 352, 384, 416]$ and v) Dropout rate - $[0.1]$. Convolutional filters and Dropout rate corresponds to the default values for ResNet. Meta-features include the value of $40$ corresponding to the number of meta-features used by \gls{FFORMA}.

For training, hyper-parameter values were selected from: i) Learning Rate - $[1e^{-3}, 1e^{-4}, 1e^{-5}]$ ii) Batch Size - $[92]$ iii) Epochs - $[150]$ iv) Early Stopping Patience - $[20]$ and v) Validation Set - $[10\%]$ of Training Set. The batch size was selected based on memory availability on the GPU. Epochs and patience were hand-tuned by evaluating training validation curves.

\subsection{Analysis}

The forecast horizon lengths for each frequency are six steps ahead for yearly data, eight steps ahead for quarterly data, $18$ steps ahead for monthly data, $13$ steps ahead for weekly data, $14$ steps ahead for daily data, and $48$ forecasts for hourly data. The M4 Competition uses \gls{OWA} as a metric for comparison. The \gls{OWA}, combines \gls{sMAPE} \cite{makridakis1993accuracy} and the mean absolute scaled error \gls{MASE} \cite{hyndman2006another}. The denominator and scaling factor of the \gls{MASE} formula are the in-sample mean absolute error from one-step-ahead predictions of the Naïve model $2$, and $n$ is the number of data points.  

\subsection{Tools and Libraries}

All algorithms were implemented in Python and ran on a mixture of Intel and AMD chips using NVIDIA GPUs. The libraries include a mix of TensorFlow, Scikit Learn and LightGBM. The source code of our experiments is available on GitHub \footnote{https://github.com/Pieter-Cawood/FFORMA-ESRNN}.

\section{Results and Discussion}

\begin{table}
\centering
\caption{Mean \gls{OWA} errors on the M4 test set and Schulze Rank~$\downarrow$~\cite{schulze2011new}. Bold is best, and grey is second best.} \label{tab::owa_results}
\resizebox{\columnwidth}{!}{
\begin{tabular}{l|rrrrrr|r}
{}&
\makecell[r]{H\\(0.4K)}&
\makecell[r]{D\\(4.2K)}&
\makecell[r]{W\\(0.4K)}&
\makecell[r]{M\\(48K)}&
\makecell[r]{Y\\(23K)}&
\makecell[r]{Q\\(24K)} &
\makecell[r]{$\downarrow$} \\ 
\bottomrule
\toprule
{AVG}                    &{0.847}   &{0.985}   &{0.860}   &{0.863}   &{0.804}   &{0.856}   & 8 \\
{ES-RNN}                 &{0.440}   &{1.046}   &{0.864}   & {0.836}  &{0.778}   &{0.847}   & 7 \\
{FFORMS}                 &\hb{0.423}&{0.981}   &{0.740}   &{0.817}   &{0.752}   &{0.830}   & 4 \\
{N-BEATS$\dagger$}       &{0.464}   &\hb{0.974}&\hb{0.703}&{0.819}   &{0.758}   &\ha{0.800}& 4 \\
{FFORMA-N}               &{0.428}   &{0.979}   &{0.718}   &{0.813}   &{0.746}   &{0.828}   & 3 \\
{FFORMA}                 &\ha{0.415}&{0.983}   &{0.725}   &\ha{0.800}&\hb{0.732}&{0.816}   & \hb{2} \\
{DeFORMA$\ddagger$}      &\hb{0.423}&\ha{0.972}&\ha{0.700}&\hb{0.802}&\ha{0.729}&\hb{0.810}& \ha{1} \\
\bottomrule
\multicolumn{7}{l}{$\dagger$ transcribed for comparison~\cite{oreshkin2019n, zeng2021topological}.} \\
\multicolumn{7}{l}{$\ddagger$ proposed methods.}
\end{tabular}
}
\end{table}

The results presented in this section shed light on the comparison between using representation learning and hand-engineered features for forecast model fusion. The study aimed to evaluate if using representation learning could improve on \gls{FFORMA} and how the proposed method might compare to other \gls{SOTA} results when considering mean and median \gls{OWA}. Table~\ref{tab::owa_results} shows the mean \gls{OWA} for comparing the selected fusion methods. The table also shows the compared methods ranked by the Schulze rank. The grey backgrounds are the two best performers, with the top result in bold. The median results are presented alongside the Schulze rank across all the different seasonalities.

The results support the claim that \gls{DeFORMA} outperforms the original \gls{FFORMA}. Both \gls{DeFORMA} and \gls{FFORMA} improve on \gls{ES-RNN} and \gls{N-BEATS}. Interestingly in Table~\ref{tab::owa_results} where \gls{FFORMA-N} had previously been the \gls{SOTA} for Daily (D) and Weekly (W), the analysis shows that \gls{DeFORMA} now outperforms the others. This result is surprising since representation learning is expected to be data-hungry. The suspicion is that the lower data requirements would allow \gls{FFORMA} to dominate in these areas. When considering the Schulze ranking, the results support the notion that \gls{DeFORMA} is the overall best-performing model.

When considering the original \gls{FFORMA}, it is clear from \textcite{cawood2022evaluating} that late meta-learning using heterogeneous base-learners is a promising approach to model fusion. However, the same study observed that \gls{FFORMA-N} and even \gls{AVG} could outperform some smaller subsets. \gls{N-BEATS} and \gls{ES-RNN} are also competitive for specific subsets of data, probably due to the representation learning they can affect. \gls{DeFORMA} combines the benefits of representation learning from \gls{N-BEATS} and \gls{ES-RNN} together with the output layer of \gls{FFORMA-N} to allow for representation learning in a late meta-learning model fusion algorithm. The results are conclusive as to the efficacy of this approach.

\subsection{Ablations}

\begin{table}
\centering
\caption{Median \gls{OWA} errors on the M4 test set. Bold is best, and grey is second best.} \label{tab::owa_results_median}
\resizebox{\columnwidth}{!}{
\begin{tabular}{l|rrrrrr}
{}&
\makecell[r]{H\\(0.4K)}&
\makecell[r]{D\\(4.2K)}&
\makecell[r]{W\\(0.4K)}&
\makecell[r]{M\\(48K)}&
\makecell[r]{Y\\(23K)}&
\makecell[r]{Q\\(24K)} \\
\bottomrule
\toprule
{FFORMA}              &\ha{0.318}&{0.723}    &{0.529}    &\ha{0.602} &\ha{0.491} &{0.580}     \\
{FFORMA-N}            &\hb{0.326}&{0.720}    &{0.539}    &{0.614}    &{0.508}    &{0.594}     \\
{DeFORMA$^\ddagger$}  &{0.330}   &\hb{0.711}    &\hb{0.525} &{0.603}    &\ha{0.491} &\hb{0.576}  \\
\midrule
{DeFORMA(224)}        &{0.342}   &\ha{0.708}    &\ha{0.517} &\ha{0.602} &   {0.499} &\ha{0.575}  \\
\bottomrule
\multicolumn{7}{l}{$\ddagger$ proposed methods.}
\end{tabular}
}
\end{table}

The results in Table~\ref{tab::owa_results_median} show the median OWA for \gls{DeFORMA}(224). In this instance, the hyper-parameter for time series length was left at the original $224$ of ResNet-18 and all halvings and filters as per the original architecture. The results show the robustness of \gls{DeFORMA} to these hyper-parameters. 

The experiments were also completed using data augmentation, which led to no improvement and VGG-11, which also led to no improvements in the results. The hyper-parameter selection for the number of meta-features learned impacted the results. Finally, excluding both or one of the Temporal Heads led to a significant drop in performance across all experiments.

\section{Conclusion}

The study sought to compare a representation learning late meta-learning model fusion approach to several other  methods and to provide a complete taxonomy for the discussion of forecast model fusion. The results demonstrate that the proposed method, \gls{DeFORMA}, is, in fact, \gls{SOTA} for the M4 dataset. These results are noteworthy since using representing learning overcomes the need for hand-engineering features. It is worth considering that the results are applied in a univariate setting and, as such, will need further work to confirm them in the hierarchical and multivariate settings. Given that representation learning is effective, the next question is whether transfer learning can be applied in this setting across seasonalities.

\renewcommand*{\bibfont}{\normalfont\footnotesize}
\printbibliography

\end{document}